\documentclass[letterpaper, 10 pt, conference]{ieeeconf}  
\usepackage{times}
\usepackage{epsfig}
\usepackage{graphicx}
\usepackage{amsmath}
\usepackage{amssymb}
\usepackage{multirow}
\usepackage{soul}
\usepackage{booktabs}
\usepackage[pagebackref=true,breaklinks=true,colorlinks,bookmarks=false]{hyperref}

\IEEEoverridecommandlockouts                              

\title{\LARGE \bf
ViNet: Pushing the limits of Visual Modality for \\ Audio-Visual Saliency Prediction
}

\author{Samyak Jain$^{1}$,  Pradeep Yarlagadda$^{1}$,   Shreyank Jyoti$^{1}$,  Shyamgopal Karthik$^{1}$,  \\ Ramanathan Subramanian$^{2}$,  Vineet Gandhi$^{1}$ 
\thanks{$^{1}$ CVIT, KCIS, International Institute for Information Technology, Hyderabad
        {\tt\small samyak.j@research.iiit.ac.in}}%
 \thanks{$^{2}$University of Canberra}%
}

\begin{document}
\def\x{{\mathbf x}}
\def\L{{\cal L}}
\def\eg{\textit{e.g.}}
\def\ie{\textit{i.e. }}
\def\Eg{\textit{E.g.}}
\def\etal{\textit{et al.}}
\def\etc{\textit{etc}}

\maketitle
\thispagestyle{empty}
\pagestyle{empty}

\begin{abstract}
We propose the \textbf{ViNet} architecture for audio-visual saliency prediction. ViNet is a fully convolutional encoder-decoder architecture. The encoder uses visual features from a network trained for action recognition, and the decoder infers a saliency map via trilinear interpolation and 3D convolutions, combining features from multiple hierarchies. The overall architecture of ViNet is conceptually simple; it is causal and runs in real-time (60 fps). ViNet does not use audio as input and still outperforms the state-of-the-art audio-visual saliency prediction models on nine different datasets (three visual-only and six audio-visual datasets). ViNet also surpasses human performance on the CC, SIM and AUC metrics for the \emph{AVE} dataset, and to our knowledge, it is the first model to do so. We also explore a variation of ViNet architecture by augmenting audio features into the decoder. To our surprise, upon sufficient training, the network becomes agnostic to the input audio and provides the same output irrespective of the input. Interestingly, we also observe similar behaviour in the previous state-of-the-art models~\cite{tsiami2020stavis} for audio-visual saliency prediction. Our findings contrast with previous works on deep learning-based audio-visual saliency prediction, suggesting a clear avenue for future explorations incorporating audio in a more effective manner. The code and pre-trained models are available at \url{https://github.com/samyak0210/ViNet}.

\end{abstract}

\section{INTRODUCTION}


Video saliency prediction focuses on understanding and modeling human visual attention (HVA) while viewing a dynamic scene (determining where and what people pay attention to given visual stimuli). HVA empowers primates to analyze/interpret the complex surroundings rapidly, and naturally, we would like to extend these abilities to machines/robots. For instance, a robot that orients its eyes like humans gives impressions of an intelligent behaviour~\cite{butko2008visual}. Moreover, it may allow the robot to orient towards regions of the visual scene that are likely to be relevant. Upon compiling the \emph{ground truth} regarding where viewers gaze in the scene via eye-tracking hardware, saliency prediction (SP) aims to mimic HVA given a novel video computationally. Previous works have shown that SP is valuable in a variety of applications like human-robot interaction~\cite{ferreira2014attentional, schillaci2013evaluating, chang2019salgaze, mavani2017facial}, robotic camera control~\cite{butko2008visual}, motion tracking~\cite{zhang2009visual}, stream compression~\cite{hadizadeh2013saliency, gupta2013visual}, video captioning \cite{nguyen2013static}, automated cinematic editing~\cite{moorthy2020gazed},~\etc.









Video SP models primarily employ visual information to predict gaze. Larger datasets like \emph{DHF1K}~\cite{wang2018revisiting} discard audio during ground truth collection, and ask users to look at \emph{silent} videos. End-to-end deep saliency models are then trained using only visual information. State-of-the-art video SP models largely depend on Long Short-Term Memory (LSTM) networks to encode temporal dependencies~\cite{droste2020unified, wu2020salsac, linardos2019simple}. These models build on image-based saliency and aggregate frame-wise prediction using an LSTM. Since both spatial decoding and temporal aggregation are performed separately, LSTM models cannot collectively leverage Spatio-temporal information, shown to be beneficial for video SP \cite{min2019tased}.


\begin{figure}[t]
\centering
\includegraphics[width=0.47\textwidth]{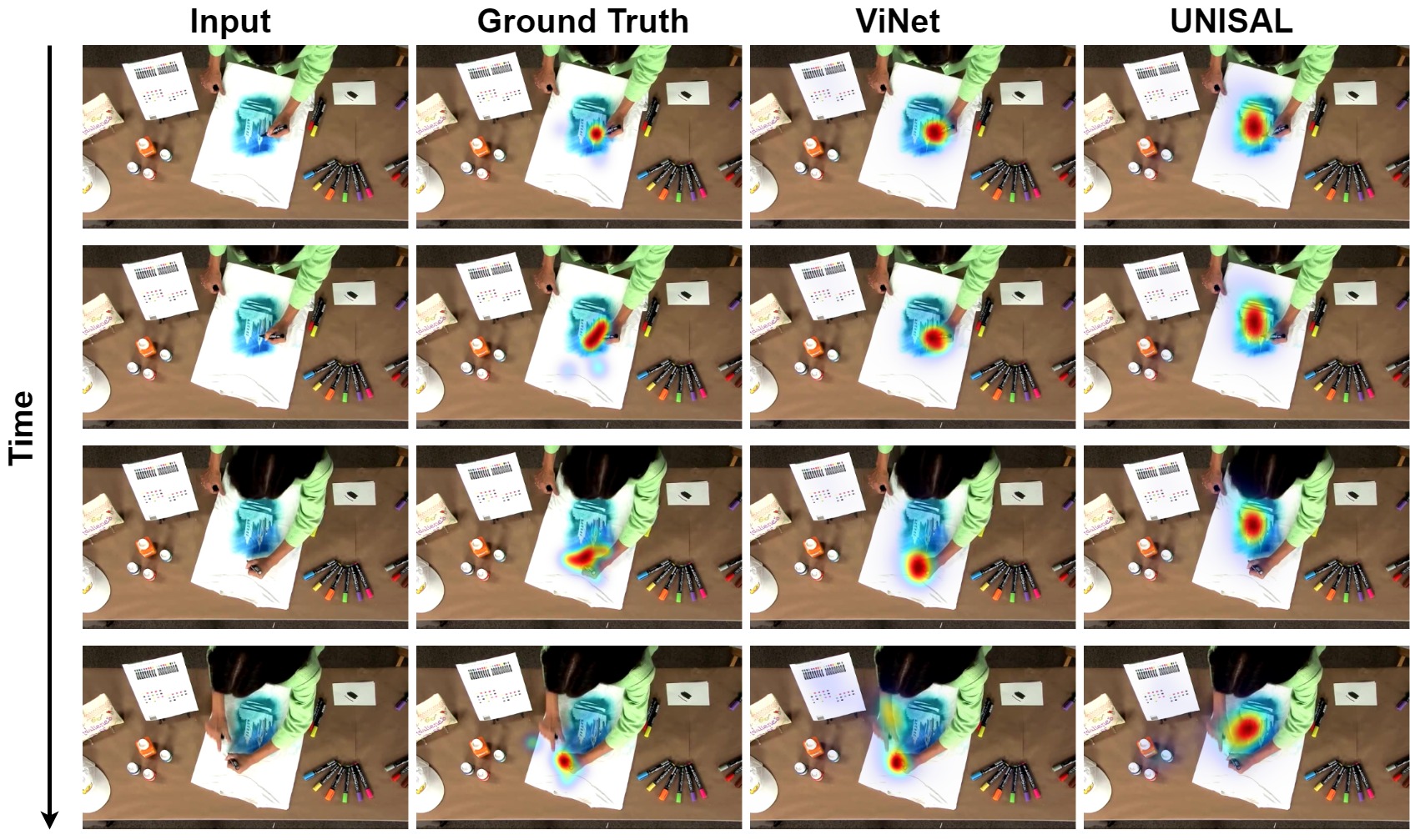}
\caption{{The core of our approach is a strong visual-only model ViNet. Here, we compare ViNet (third column) with state-of-the-art UNISAL model~\cite{droste2020unified} (fourth column). Note that ViNet better captures the action, while UNISAL focuses on objectness. In this example, ViNet focuses on the region being drawn, whereas UNISAL focuses on the completed portion. Best viewed in color and under zoom.}}\vspace{-.1cm}
\label{fig:DHF1K_results}
\end{figure}

\begin{figure*}[t]
\centering
\includegraphics[width=0.9\textwidth]{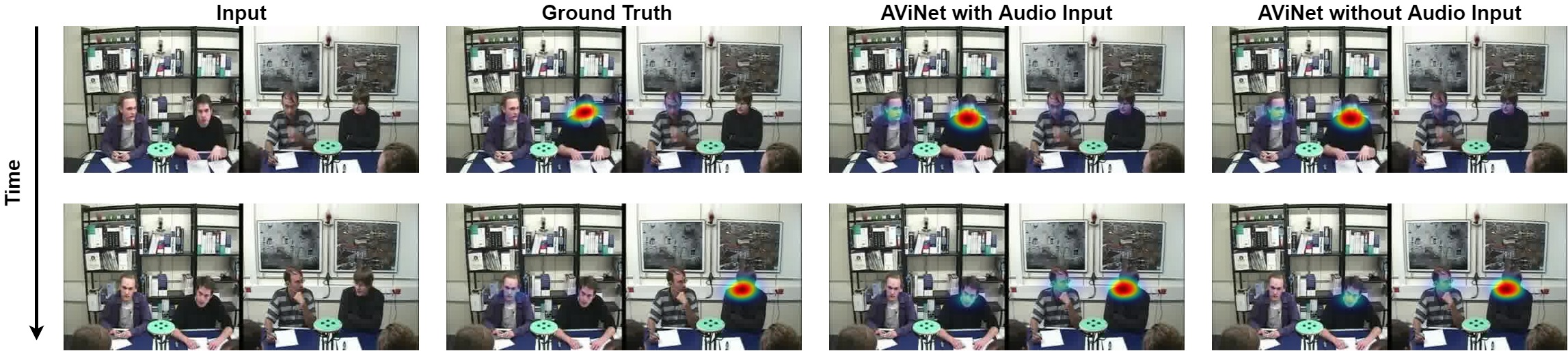}
\caption{{Sample frames from \emph{Coutrot-2} database with the corresponding ground-truth. The predicted saliency maps of AViNet with and without passing audio input turn out to be the same.}}\vspace{-.1cm}
\label{fig:avinet_results}
\end{figure*}

To this end, we propose a novel fully convolutional encoder-decoder architecture called ViNet for visual saliency detection. ViNet takes a set of frames as input and predicts a saliency map for the last frame. Following the methodology adopted in~\cite{min2019tased}, it then employs a sliding window approach to predict saliency for the entire video. ViNet takes features learned from an action recognition network from multiple hierarchies, fuses them in a UNet~\cite{ronneberger2015u} like fashion, and outputs a saliency map using trilinear interpolations and 3D convolutions. The strength of ViNet is that it only comprises commonly used components, resulting in a minimal and conceptually simple model which is easy to train and interpret. ViNet is causal, runs in real-time, and surpasses the state-of-the-art on three popular vision-only saliency prediction datasets (a motivating example is illustrated in Fig.~\ref{fig:DHF1K_results}). At the time of submission, ViNet is also the top-ranked model on the private test-set of \emph{DHF1K}, the most diverse video saliency prediction benchmark. Interestingly, ViNet also achieves state-of-the-art results on six audio-visual saliency datasets without using any audio information.

More fundamentally, discarding audio information contrasts with our real-life behaviour, where we simultaneously perceive visual and audio modalities. Cognitive studies confirm that auditory and visual cues are correlated and jointly contribute to human attention~\cite{van2008audiovisual}. Coutrot \etal ~\cite{coutrot2012influence} collect the human gaze on the same set of videos with and without the original soundtrack and observe that the soundtrack significantly affects the attention models in human perception, even when using a monophonic stimuli~\cite{coutrot2012influence}. Consequently, recent efforts have explored multi-modal (audio-visual) video SP~\cite{tavakoli2019dave,tsiami2020stavis}, and claim audio as a strong  cue for SP. 



Consequently, we experiment with an audio-visual saliency prediction model obtained by augmenting ViNet with an audio branch. The resulting architecture called AViNet is end-to-end trainable and uses pre-trained audio features from SoundNet~\cite{aytar2016soundnet}. We explore two fusion strategies, similar to~\cite{tsiami2020stavis, tavakoli2019dave} \ie \ simple concatenation and bilinear fusion. We observe that when compared to ViNet, AViNet gives nil or marginal improvements on most audio-visual SP datasets. Our results suggest that current audio-visual saliency models~\cite{tavakoli2019dave,tsiami2020stavis} are not optimal on the visual modality. Furthermore, when we dig deeper, we find out that the audio-visual network learns to ignore the audio signal entirely and gives the same result even while sending a zero vector as audio or by sending an unrelated random audio file (Fig. ~\ref{fig:avinet_results}). Surprisingly, we observe the same behaviour with STAViS~\cite{tsiami2020stavis}, the current state-of-the-art audio-visual saliency prediction model. Our findings contrast to the prevalent claims that the audio acts as a strong cue for SP. Overall, we make the following research contributions:





\begin{itemize}
    \item We propose a novel visual-only architecture called ViNet for video saliency detection. Our model uses commonly known deep learning components/ideas, and the contributions are in their efficient amalgamation. We back the proposed architecture with thorough ablation studies.
    \item We present a comprehensive analysis on ten different datasets (three visual and seven audio-visual datasets). Our model achieves solid performance gains over the current state-of-the-art.
     \item  We carefully explore the role of audio and find that the visual-only model almost recovers the underlying performance. Furthermore, the strategies mentioned in existing literature end up learning a prediction model agnostic to audio. This motivates the need for exploring novel architectures for audio-visual fusion for SP and possibly carefully curating datasets where audio plays a significant role. 
\end{itemize}

\begin{figure*}[]
\includegraphics[width=0.95\textwidth]{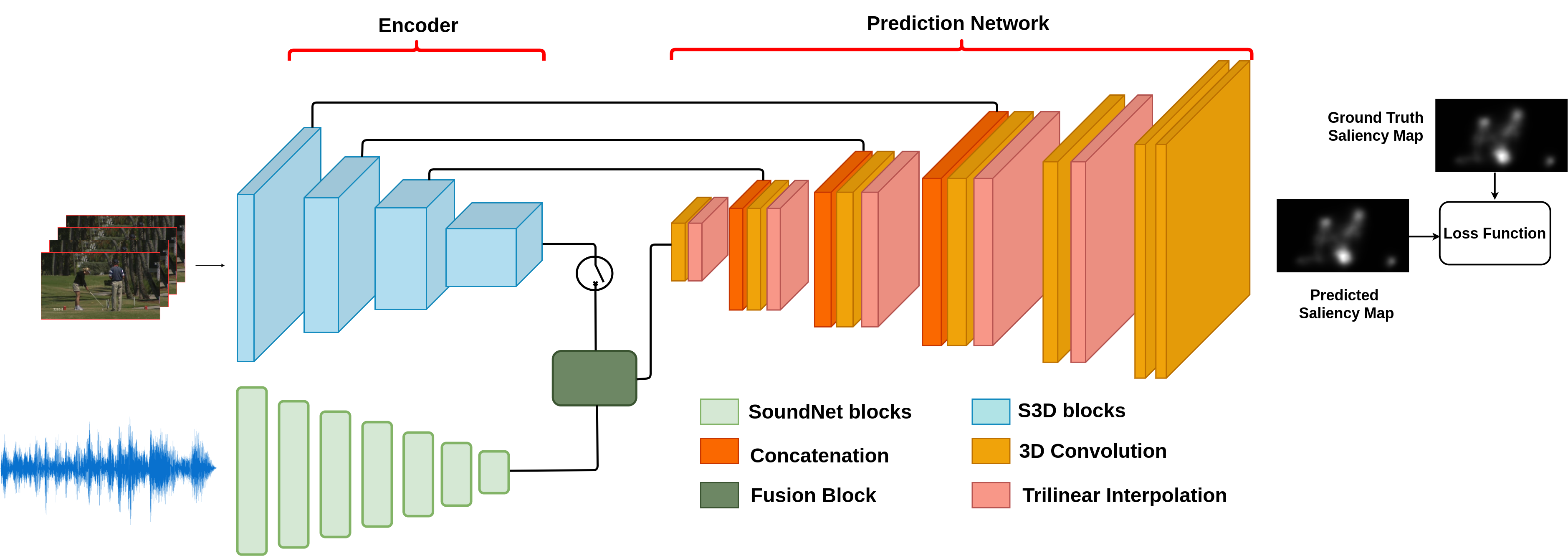}
\caption{AViNet Architecture overview. Removing the audio branch, the resulting architecture is ViNet. }
\label{fig:AViNet}
\end{figure*}

\section{Related Works}

\subsection{Video Saliency}

The recent landscape in video saliency prediction is dominated by the end-to-end trainable deep networks. The availability of large datasets like \emph{Hollywood-2}~\cite{marszalek2009actions} and \emph{DHF1K}~\cite{wang2018revisiting} have been instrumental in this progress. \emph{Hollywood-2} is the largest dataset, however, its content is limited to human actions and movie scenes. \emph{DHF1K} is considered the most diverse and challenging dataset for saliency detection.

Majority of the recent approaches rely on an LSTM based architecture for sequential fixation prediction over successive frames. Wang \etal~\cite{wang2018revisiting} combine frame-level image features using a ConvLSTM. SalEMA~\cite{linardos2019simple} model recurrence using a temporal exponential moving average (EMA) operation over the convolutional layer. They show such a simple moving average-based approach matches the performance achieved using a ConvLSTM. SALSAC~\cite{wu2020salsac} adds further complexity to basic ConvLSTM architecture through a shuffled multi-level attention module and a frame correlation module. STRA-Net~\cite{chen2020video} learns an alignment module, and then aligned frames are sent into a Bi-ConvLSTM. ~\cite{jiang2017predicting} propose a novel construction of LSTM (2C-LSTM) with two sub-networks to focus on objectness and motion, respectively. UNISAL~\cite{droste2020unified} is a unified image and video saliency prediction model that uses MobileNet to extract spatial features and LSTMs for encoding temporal information. The method heavily relies on domain adaptive prior maps (different prior maps for image and video domains), domain adaptive batch-normalization, etc. Several of these video saliency prediction architectures~\cite{droste2020unified,wang2018revisiting} borrow and extend ideas (hierarchical features, transfer learning, multi-branch architectures, etc.) from the models trained for static image saliency prediction~\cite{jiang2015salicon,reddy2020tidying}.  

3D convolutional architectures have also been explored for the task. These methods typically rely on action detection networks as their backbone. TASED-Net ~\cite{min2019tased} uses S3D as an encoder to extract spatial features while jointly aggregating all the temporal information in order to produce a single full-resolution prediction map. They use transposed convolution layers with auxiliary pooling ( a variation of max-unpooling layers) for spatial upscaling in the decoder. Bellitto~\etal~\cite{bellitto2020video} use multiple decoders for features encoded at different levels to obtain multiple saliency instances that are finally combined to obtain final output saliency maps. ~\cite{bellitto2020video} is inspired by the DVA image saliency model~\cite{wang2017deep}. A combination of 3D convolutions and recurrent architecture has also been explored~\cite{bazzani2016recurrent}. STSConvNet~\cite{bak2017spatio} explicitly computes optical flow and fuses the optical with the visual features into two-stream convolutional architecture.

In contrast, our ViNet method is a straightforward encoder-decoder architecture exploiting basic ideas of spatial hierarchy, feature concatenation, skip connections, trilinear upsampling, and 3D convolutions. It uses pretrained features from a network trained for action recognition as a backbone and is void of any explicit inputs like optical flow or any extra modules for detecting objectness, motion, attention, etc. 









\subsection{Audio-Video Saliency}

Research in cognitive neuroscience has led to interesting findings about audiovisual integration. If you ever watched a ventriloquist in action, you would agree how they trick our visual stimuli to guide the perceived location of the sound (and where we look at). Ventriloquist turns to face the puppet, they attend the puppet, use a different voice for the puppet, and make it seem that it is the puppet that is talking (although the sound is being generated from their stomach). McGurk effect~\cite{mcgurk1976hearing}, pip and pop effect~\cite{van2008pip}, unity assumptions~\cite{vatakis2007crossmodal} are other examples of how we jointly integrate and perceive visual and audio modalities. Coutrot \etal~\cite{coutrot2012influence, coutrot2014saliency, coutrot2016multimodal} present some interesting studies on the influence of soundtrack on eye movements during video exploration.  



Application-specific attempts have been made for visual saliency and audio localization~\cite{ruesch2008multimodal,schauerte2011multimodal,chen2014audio}. The fusion of handcrafted attention models and pre-trained deep image-level models using canonical correlation analysis has been explored~\cite{min2016fixation,min2020multimodal}. However, only a couple of attempts have been made towards an end-to-end deep learning-based audio-visual saliency fixation prediction. Tavakoli et al.~\cite{tavakoli2019dave} trains two independent networks for the two modalities (audio and visual data), and their outputs are simply concatenated as a late fusion scheme. They use 3DResNet as the backbone for both modalities. STAViS~\cite{tsiami2020stavis} extends the SUSiNet~\cite{koutras2019susinet} visual saliency model and investigates three different ways to fuse the audio modality.  

Significant efforts have been made in the direction of self-supervised learning and representation learning exploiting audio-visual data. SoundNet~\cite{aytar2016soundnet} leverage the natural synchronization between vision and sound to learn an acoustic representation. They use a student-teacher training procedure to transfer discriminative visual knowledge (large-scale visual recognition) into the sound modality. On similar lines, audiovisual correspondence has been exploited for the task of cross-modal retrieval~\cite{arandjelovic2018objects}, sound classification~\cite{aytar2016soundnet, arandjelovic2017look}, sound localization in images~\cite{arandjelovic2018objects, senocak2018learning}, scene analysis~\cite{owens2018audio}, temporal event localization~\cite{tian2018audio} etc.

\section{Proposed Architecture}

We propose an end-to-end architecture visual-only model called ViNet. It is a fully 3D-convolutional encoder-decoder architecture that predicts the saliency for the last frame of the corresponding set of sequential frames. Then we present an audio-visual saliency detection model called AViNet that fuses the visual features from ViNet and audio features from SoundNet. Fig.~\ref{fig:AViNet} displays an overview of the architecture. 

\subsection{Backbone}
The architecture uses the S3D network~\cite{xie2018rethinking} as the video encoder. We use the model pre-trained on the Kinetics dataset which is an action-recognition dataset. We use S3D since it consists of 3D convolutional layers which efficiently encodes the spatio-temporal information. Moreover, it is light-weight and pre-trained on a large dataset, making it fast and effective for transfer-learning. It consists of 4 convolutional blocks base1, base2, base3 and base4 that provides outputs $X_1$,$X_2$,$X_3$ and $X_4$ in different spatial and temporal scales. $X_1$,$X_2$ and $X_3$ are referred as multi-level features that are extracted at three-levels of hierarchy. The input to the encoder is a video clip $x_{clip}\in R^{3\times T_{0} \times H_{0} \times W_{0}}$, where $T_{0}$ is 32. It generates a lower-resolution activation map $X_4\in R^{C\times T \times H \times W}$, where $C=1024$ and $T, H, W=\frac{T_{0}}{8}, \frac{H_{0}}{32}, \frac{W_{0}}{32}$.

For audio representation, we employ SoundNet~\cite{aytar2016soundnet}, which is trained for audio/sound based scene classification. We pre-process the audio data similar to the STAViS~\cite{tsiami2020stavis} (section 3.2). The sound module takes 1D pre-processed audio feature as input, $y_{audio}\in R^{1\times\hat{T}\times1}$ and outputs audio features $A\in R^{1024\times3\times1}$.  


\subsection{Audio-Visual Fusion}
Inspired by the recent works on audio-visual saliency prediction~\cite{tsiami2020stavis,tavakoli2019dave}, we explore two types of fusion techniques. First is a simple concatenation of encoded audio and video features which was used in ~\cite{tavakoli2019dave}. We repeat the audio features to match the dimensions of visual features and combine them across the channel dimension. Then we apply $1\times1$ Convolution to reduce the number of channels.

Secondly, we applied bilinear fusion which has been used in \cite{tsiami2020stavis}. The visual features are first passed through Max Pool to reduce the spatial and temporal dimension and then collapsed to represent it as a vector $x_{1}\in R^{1024\times x_{0}}$. Similarly, the audio features are collapsed as a vector $x_{2}\in R^{1024\times y_{0}}$. The bilinear fusion is defined as 
\begin{equation}
    y = x_{1}^TAx_{2} + b
\end{equation}
where $A\in R^{x_{0}\times x \times y_{0}}$ and $b\in R^{x\times 1}$ are parameters and $x$ is the desired output dimension. 
\begin{table}[]
    \caption{Validation results on varying clip size for training ViNet on \emph{DHF1K}.}
    \begin{center}
    
    \begin{tabular}{|c|ccc|}
    \hline
       \textbf{Clip Size ($T$)} & CC  & SIM & NSS \\
    \hline\hline
    \textbf{8}           &0.4978	&0.363	&2.8221 \\
    \textbf{16}           &0.5112	&0.378	&2.9067 \\
    \textbf{32}          &0.5212	&\textbf{0.3881}	&\textbf{2.9565} \\
    \textbf{48}          &\textbf{0.5231}	&0.3807	&2.9477 \\
    \hline
    \end{tabular}
    \end{center}
    
    \label{tab:clip_length}
\end{table}

\begin{table}[]
    \caption{Validation results of ViNet with and without hierarchy on \emph{DHF1K}.}
    \begin{center}
        
    \begin{tabular}{|c|ccc|}
    \hline
       \textbf{Model Architecture} & CC  & SIM & NSS \\
    \hline\hline
    \textbf{Without Hierarchy}           &0.5002	&0.361	&2.7371 \\
    \textbf{With Hierarchy}           &0.5212	&0.3881	&2.9565 \\
    \hline
    \end{tabular}
    \end{center}
    
    \label{tab:hierarchy}
\end{table}

\begin{table*}[t]
    \caption{Comparison results on the \emph{DHF1K}, \emph{Hollywood-2} and \emph{UCF-Sports} test sets. The best scores are shown in red and second best scores in blue. FPS values are taken from the leaderboard and are computed on different hardware, hence, a direct comparison cannot be made. ViNet is evaluated on a GTX 1080 Ti GPU and runs at a speed adequate for most real-time applications.}
     \footnotesize
     \begin{center}

\resizebox{\textwidth}{!}{
    \begin{tabular}{|c|c|ccccc|ccccc|ccccc|}
    \hline
     & \multicolumn{1}{c|}{\emph{FPS}} & \multicolumn{5}{c|}{\emph{DHF1K}} &  \multicolumn{5}{c|}{\emph{Hollywood-2}} &  \multicolumn{5}{c|}{\emph{UCF-Sports}} \\
        & & CC  & sAUC     & AUC    & NSS    & SIM & CC  & sAUC     & AUC    & NSS    & SIM & CC  & sAUC     & AUC    & NSS    & SIM\\
    \hline\hline
    \textbf{SALEMA}~\cite{linardos2019simple}               &100  &0.449	&0.667	&0.890	&2.57	&\textcolor{red}{0.466}	&0.613	&0.708	&0.919	&3.18	&0.487	&0.544	&0.740	&0.906	&2.63	&0.431\\
    \textbf{ACLNet}~\cite{wang2019revisiting}                &50 &0.434	&0.601	&0.890	&2.35	&0.315	&0.623	&0.757	&0.913	&3.08	&\textcolor{blue}{0.542}	&0.510	&0.744	&0.897	&2.56	&0.406 \\
    \textbf{STRA-Net}~\cite{lai2019video}               &50 &0.458	&0.663	&0.895	&2.55	&0.355	&0.662	&0.774	&0.923	&3.47	&0.536	&0.593	&0.751	&0.910	&3.01	&0.479 \\
    \textbf{SALSAC}~\cite{wu2020salsac}                 &50 &0.479	&0.697	&0.896	&2.67	&0.357	&0.670	&0.712	&0.931	&3.35	&0.529	&\textcolor{blue}{0.671}	&\textcolor{blue}{0.806}	&\textcolor{red}{0.926}	&3.52	&\textcolor{red}{0.534}\\
    \textbf{TASED-Net}~\cite{min2019tased}               &16.6 &0.470	&\textcolor{blue}{0.712}	&0.895	&2.66	&0.361	&0.64	&0.768	&0.918	&3.30	&0.507	&0.582	&0.752	&0.899	&2.92	&0.469\\
    \textbf{UNISAL}~\cite{droste2020unified}                 &111 &\textcolor{blue}{0.490}	&0.691	&\textcolor{blue}{0.901}	&\textcolor{blue}{2.77}	&\textcolor{blue}{0.390}	&\textcolor{blue}{0.673}	&\textcolor{blue}{0.795}	&\textcolor{red}{0.934}	&\textcolor{red}{3.90}	&\textcolor{blue}{0.542}	&0.644	&0.775	&0.918	&\textcolor{blue}{3.38}	&\textcolor{blue}{0.523}\\
    \hline
    \textbf{ViNet}                  	& 62.5 &\textcolor{red}{0.510}	&\textcolor{red}{0.728}	&\textcolor{red}{0.908}	&\textcolor{red}{2.87}	&0.381	&\textcolor{red}{0.693}	&\textcolor{red}{0.813}	&\textcolor{blue}{0.930}	&\textcolor{blue}{3.73}	&\textcolor{red}{0.550}	&\textcolor{red}{0.673} &	\textcolor{red}{0.810}	&\textcolor{blue}{0.924}	&\textcolor{red}{3.62}	&0.522\\
    \hline
    \end{tabular}}
    \end{center}
    
    \label{table:test_results}
\end{table*}


\subsection{Prediction Network}

The Prediction Network consists of 6 decoding layers consisting of 3D convolutional and upsampling layers. For ViNet, the input to the Prediction Network is the $X_4$ features from the Backbone and $X_3$,$X_2$, and $X_1$ are passed in using skip connections, respectively. In the case of AViNet, the audio features are fused with $X_4$ and then sent to the decoder (skip connections are made similarly). 


\subsection{Evaluation}
Both ViNet and AViNet follow a sliding window approach to generate a saliency map for all frames in the video. Given a window size of $T$ frames, we predict saliency map $S_{t}$ at time step $t$ by taking $F_{t-T+1},...F_{t}$ sequence of frames as input. To enable prediction in the first $T$ frames, we simply repeat the first frame of the video at the start. A single inference of ViNet takes around 0.016 seconds (62.5 fps) to generate a saliency map, with $T$ = 32 frames.



\section{Experiments and Results}
\subsection{Datasets}
\subsubsection{Visual Datasets}
The three most popular visual-only datasets in video saliency are \emph{DHF1K}, \emph{Hollywood-2}, and \emph{UCF-Sports}~\cite{rodriguez2008action}. We carry out the tests and comparisons on these three datasets. 

\emph{DHF1K}~\cite{wang2018revisiting} contains 1000 videos where 700 videos are for training and 100 for validation. A test set of 300 videos is also released, however, without public ground truth. All our experiments and analysis are based on this dataset since it is the most general and diverse dataset.

\emph{Hollywood-2}~\cite{marszalek2009actions} is the largest video saliency prediction dataset in terms of the number of videos, consisting of 1707 videos. The dataset is focused on human actions. The videos in this dataset are short video sequences from a set of 69 Hollywood movies, containing 12 different human action classes, ranging from answering the phone, eating, driving, running and etc. We use the standard split of 823  training videos and 884 test videos.

\emph{UCF-Sports}~\cite{rodriguez2008action} dataset consists of 150 videos focusing on human actions in sports. We use a standard split with 103 videos for training and 47 videos for testing. 

\subsubsection{Audio-Visual Datasets}
There are seven audio-visual datasets in video saliency: \emph{DIEM}, \emph{Coutrot1}, \emph{Coutrot2}, \emph{AVAD}, \emph{ETMD}, \emph{SumMe}, and \emph{AVE} dataset. We carry out the tests and comparisons on all these seven datasets. 

\emph{DIEM}~\cite{mital2011clustering} consists of 81 movie clips of varying genres. They sourced from publicly accessible repositories, including advertisements, documentaries, game trailers, movie trailers, music videos, news clips, and time-lapse footage. It consists of 64 training videos and 17 test videos.

\emph{Coutrot} databases~\cite{coutrot2014saliency,coutrot2016multimodal} are split into \emph{Coutrot1} and \emph{Coutrot2}. \emph{Coutrot1} contains 60 clips with dynamic natural scenes split into four visual categories: one/several moving objects, landscapes, and faces. \emph{Coutrot2} contains 15 clips of 4 persons in a meeting and the corresponding eye-tracking data from 40 persons. 

\emph{AVAD} dataset~\cite{min2016fixation} contains 45 short clips of 5-10 sec duration with several audio-visual scenes, e.g., dancing, guitar playing, birds singing, etc. 

\emph{ETMD} dataset~\cite{koutras2015perceptually} contains 12 videos from six different hollywood movies.

\emph{SumMe} dataset~\cite{gygli2014creating} contains 25 unstructured videos, \ie, mostly user-made videos and their corresponding multiple-human created summaries, which were acquired in a controlled psychological experiment.

\emph{AVE} dataset~\cite{tavakoli2019dave} consists of 150 hand-picked video sequences from \emph{DIEM}, \emph{Coutrot1} and \emph{Coutrot2} datasets. The videos are divided into three categories - Nature, Social Events, and Miscellaneous. The dataset consists of 92 training videos, 29 validation, and 29 test sequences.

\subsection{Experimental Setup}
\label{subsec:Expsetup}

\begin{table*}[t]
\caption{Comparison results on the \emph{DIEM}, \emph{Coutrot1}, \emph{Coutrot2}, \emph{AVAD}, \emph{ETMD} and \emph{SumMe} test sets.}
\footnotesize
\begin{center}

\begin{tabular}{|c|ccccc|ccccc|ccccc|}
\hline
 & \multicolumn{5}{c|}{\emph{DIEM}} & \multicolumn{5}{c|}{\emph{Coutrot1}} & \multicolumn{5}{c|}{\emph{Coutrot2}} \\
 & CC & sAUC & AUC & NSS & SIM & CC & sAUC & AUC & NSS & SIM & CC & sAUC & AUC & NSS & SIM \\ 
 \hline\hline
\textbf{ACLNet}~\cite{wang2019revisiting} & 0.522 & 0.622 & 0.869 & 2.02 & 0.427 & 0.425 & 0.542 & 0.85 & 1.92 & 0.361 & 0.448 & 0.594 & 0.926 & 3.16 & 0.322 \\
\textbf{TASED-Net}~\cite{min2019tased} & 0.557 & 0.657 & 0.881 & 2.16 & 0.461 & 0.479 & 0.58 & 0.867 & 2.18 & 0.388 & 0.437 & 0.611 & 0.921 & 3.17 & 0.314 \\
\textbf{STAViS}~\cite{tsiami2020stavis} & 0.579 & 0.674 & 0.883 & 2.26 & 0.482 & 0.472 & 0.584 & 0.868 & 2.11 & 0.393 & 0.734 & 0.71 & \textbf{0.958} & 5.28 & \textbf{0.511} \\
\hline
\textbf{ViNet(NF)} & 0.571 & 0.695 & 0.886 & 2.28 & 0.468 & 0.509 & 0.619 & 0.875 & 2.46 & 0.406 & 0.645 & 0.72 & 0.949 & 5.11 & 0.419 \\
\textbf{ViNet} & 0.626 & \textbf{0.723} & 0.898 & 2.47 & 0.483 & 0.551 & 0.633 & 0.886 & 2.68 & 0.423 & 0.724 & 0.739 & 0.95 & 5.61 & 0.466 \\
\textbf{AViNet(B)} & \textbf{0.632} & 0.719 & \textbf{0.899} & \textbf{2.53} & \textbf{0.498} & \textbf{0.56} & 0.635 & \textbf{0.889} & \textbf{2.73} & 0.425 & \textbf{0.754} & \textbf{0.742} & 0.951 & \textbf{5.95} & 0.493 \\ 
\textbf{AViNet(C)} & 0.631&	0.720&	0.897&	2.50&	0.497 &0.556&	\textbf{0.636}&	0.887&	2.68&	\textbf{0.426} & 0.753&	0.743&	0.951&	5.81&	0.486 \\ \hline
\end{tabular}

\begin{tabular}{|c|ccccc|ccccc|ccccc|}
\hline
 & \multicolumn{5}{c|}{\emph{AVAD}} & \multicolumn{5}{c|}{\emph{ETMD}} & \multicolumn{5}{c|}{\emph{SumMe}} \\
 & CC & sAUC & AUC & NSS & SIM & CC & sAUC & AUC & NSS & SIM & CC & sAUC & AUC & NSS & SIM \\ 
 \hline\hline
\textbf{ACL-Net}~\cite{wang2019revisiting} & 0.58 & 0.56 & 0.905 & 3.17 & 0.446 & 0.477 & 0.675 & 0.915 & 2.36 & 0.329 & 0.379 & 0.609 & 0.868 & 1.79 & 0.296 \\
\textbf{TASED-Net}~\cite{min2019tased} & 0.601 & 0.589 & 0.914 & 3.16 & 0.439 & 0.509 & 0.711 & 0.916 & 2.63 & 0.366 & 0.428 & 0.657 & 0.884 & 2.1 & 0.333 \\
\textbf{STAViS}~\cite{tsiami2020stavis} & 0.608 & 0.593 & 0.919 & 3.18 & 0.457 & 0.569 & 0.731 & \textbf{0.931} & 2.94 & \textbf{0.425} & 0.422 & 0.656 & 0.888 & 2.04 & 0.337 \\ \hline
\textbf{ViNet (NF)} & 0.665 & 0.651 & 0.923 & 3.67 & 0.501 & 0.544 & 0.719 & 0.924 & 2.92 & 0.404 & 0.455 & 0.687 & 0.893 & 2.35 & \textbf{0.349} \\
\textbf{ViNet} & \textbf{0.694} & \textbf{0.663} & 0.928 & \textbf{3.82} & \textbf{0.504} & 0.569 & 0.736 & 0.928 & 3.06 & 0.409 & 0.466 & 0.696 & 0.898 & 2.40 & 0.345 \\
\textbf{AViNet(B)} & 0.674 & 0.658 & 0.927 & 3.77 & 0.491 &\textbf{0.571} & 0.733 & 0.928 & \textbf{3.08} & 0.406 & 0.463 & 0.692 & 0.897 & 2.41 & 0.343 \\
\textbf{AViNet(C)} & 0.683 & 0.661 & \textbf{0.931} & 3.74 & 0.494 & 0.566 & \textbf{0.737} & 0.928 & 3.05 & 0.404 & \textbf{0.471} & \textbf{0.699} & \textbf{0.899} & \textbf{2.42} & 0.346 \\ \hline
\end{tabular}

\end{center}

\label{table:audio_test_results1}
\end{table*}

For training ViNet, clips with $T$ consecutive frames were randomly selected from the dataset. Each frame is resized to $224\times384$ and trained with a batch size of $8$. The optimizer used is Adam, and the learning rate is set to be 1e-4. The network is initially trained on the \emph{DHF1K} dataset. The validation set of \emph{DHF1K} is used for early stopping. The trained model is then fine-tuned for \emph{Hollywood-2} and \emph{UCF-Sports} dataset using their respective training sets. The test sets of \emph{Hollywood-2} and \emph{UCF-Sports} are used for early stopping.

For our audio-visual extension AViNet, weights of ViNet pre-trained on \emph{DHF1K} are used and fine-tuned on the audio-visual datasets. For \emph{DIEM}, the standard split provided in the literature is used. For other datasets, there are no standard splits defined, so we evaluated our model on three different splits defined by~\cite{tsiami2020stavis} and report the average metric values across various splits. For evaluating on \emph{AVE} dataset, we fine-tune the model using its training set and use its validation for early stopping.


\subsection{Evaluation Metrics}
 We evaluate our method on five standard evaluations metrics~\cite{bylinskii2018different}. The CC metric computes the Pearson’s Correlation between the ground truth and the predicted maps. The Similarity metric (SIM) computes the histogram intersection. The Area Under the ROC Curve (AUC) treats saliency map as a binary classifier of fixations at various threshold values, and an ROC curve is swept out by measuring the true and false positive rates under each binary classifier. sAUC is a variation of AUC where negatives are sampled from fixation locations from other images. NSS is computed as the average normalized saliency at fixated locations.

\subsection{Loss Function}
We use the Kullback-Leibler divergence as the loss function, which is often used in saliency prediction tasks. KLDiv is an information-theoretic measure of the difference between two probability distributions:
\begin{equation}
    KLdiv(P,Q) = \sum_{i} Q_{i}\log(\epsilon + \frac{Q_{i}}{P_{i} + \epsilon} ),
\end{equation}
where $P$, $Q$ are predicted and ground truth maps respectively and $\epsilon$ is a regularization term. 

\subsection{Ablation studies}

We present ablations studies that motivated our design choices in the ViNet model. All the ablations in this section are performed with training on the \emph{DHF1K} training set and evaluation on its validation set. We examine the effects of (a) changing the clip size, (b) using multi-level features, (c) replacing upsampling with transpose convolutions, and (d) applying different concatenation techniques for fusing hierarchical features.  Table~\ref{tab:clip_length} illustrates the results on varying the clip size of the input and using clips of size 32 frames gave the best results.  Ablation results by using hierarchical features can be found in Table~\ref{tab:hierarchy}. It clearly indicates that using multi-level features adds up to the performance. We also use transpose convolution instead of trilinear upsampling to increase the spatial dimension, but CC decreased to 0.5178 from 0.5212. The multi-level features extracted from the backbone are concatenated at each decoder block. We tried two ways of concatenating features - across temporal dimension and channel dimension. We observed that they gave a similar performance; therefore, we went ahead with the former approach due to fewer trainable parameters.

\subsection{Comparison with state-of-the-art}

\begin{figure*}[t!]
\centering
\includegraphics[width=0.64\textwidth]{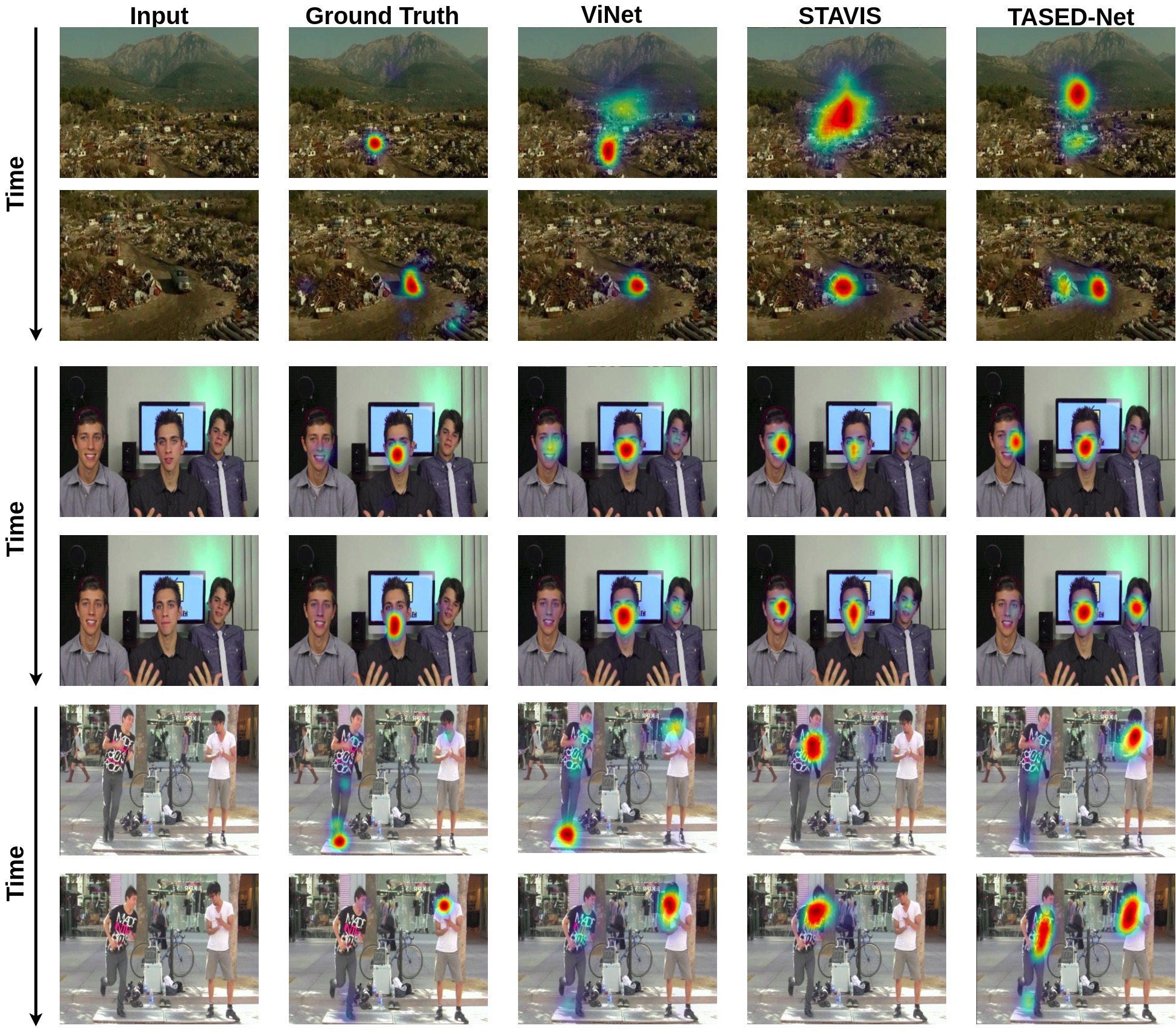}
\caption{{Sample frames from the \emph{Coutrot1} and \emph{AVAD} datasets with the corresponding ground truth, ViNet, and previous state-of-the-art STAViS and TASED-Net visual saliency maps for comparisons. ViNet is able to capture the salient region in all of these 3 examples efficiently, whereas STAViS and TASED-Net are not able to capture the salient regions accurately. Complete video for all figures is available in the supplementary material.}}
\label{fig:Comparison}
\end{figure*}

\paragraph{Visual-Only datasets}
We quantitatively compare our model with the top six state-of-the-art models on \emph{DHF1K}, \emph{Hollywood-2}, and \emph{UCF-Sports} test set. Table~\ref{table:test_results} shows the results on all three datasets in terms of CC, sAUC, AUC, NSS, and SIM metrics. We can observe that ViNet outperforms all the state-of-the-art models on the \emph{DHF1K} dataset. ViNet also achieves top results on most metrics on \emph{Hollywood-2} and \emph{UCF-Sports} datasets. At the time of the submission, ViNet is the top-performing model on the \emph{DHF1K} challenge (evaluated on the private test set)\footnote{The challenge website can be found here \url{https://mmcheng.net/videosal/}}. We show a qualitiative example in Fig. \ref{fig:Comparison} where we see that ViNet is able to produce much more accurate saliency maps as compared to TASED-Net and STAViS.


\begin{table}[t!]
\caption{Performance of various models on AVE test set categories. HI represents Human Infinite (HI) represents upper performance bound and Mean Eye Position (MEP) represents lower performance bound.}
\footnotesize
\begin{center}
\begin{tabular}{|c|c|ccccc|}
\hline
 Cat. & Model Name & CC & sAUC & AUC & NSS & SIM  \\ 
 \hline
 \multirow{4}{*}{\rotatebox[origin=c]{90}{Nature}} & HI & 0.669 & \textbf{0.762} & 0.866 & \textbf{3.32} & \textbf{0.538} \\
 & AViNet & 0.649 & 0.729 & 0.895 & 2.37 & 0.515 \\
 & ViNet & \textbf{0.680} & 0.735 & \textbf{0.900} & 2.47 & \textbf{0.538} \\
 & DAVE~\cite{tavakoli2019dave} & 0.539 & 0.723 & 0.877 & 2.27 & 0.450 \\
 & ACLNet*~\cite{wang2019revisiting} & 0.517 & 0.723 & 0.884 & 2.03 & 0.401  \\
 & MEP & 0.471 & 0.686 & 0.869 & 1.76 & 0.368 \\ \hline
 \multirow{4}{*}{\rotatebox[origin=c]{90}{Soc Ev.}} & HI & 0.655 & 0.759 & 0.855 & \textbf{3.63} & 0.516  \\
 & AViNet & \textbf{0.688} & \textbf{0.765} & \textbf{0.914} & 2.96 & 0.536 \\
 & ViNet & \textbf{0.688} & 0.760 & 0.910 & 2.88 & \textbf{0.544} \\
 & DAVE~\cite{tavakoli2019dave} & 0.545 & 0.726 & 0.885 & 2.65 & 0.442  \\
 & ACLNet*~\cite{wang2019revisiting} & 0.449 & 0.683 & 0.869 & 2.02 & 0.359  \\
 & MEP & 0.314 & 0.633 & 0.819 & 1.35 & 0.274 \\ \hline
 \multirow{4}{*}{\rotatebox[origin=c]{90}{Misc.}} & HI & 0.597 & \textbf{0.748} & 0.837 & \textbf{3.23} & 0.481 \\
 & AViNet & 0.635 & 0.730 & \textbf{0.898} & 2.42 & 0.506 \\
 & ViNet & \textbf{0.636} & 0.726 & 0.896 & 2.40 & \textbf{0.509}  \\
 & Dave~\cite{tavakoli2019dave} & 0.549 & 0.736 &  0.881 & 2.39 & 0.454  \\
 & ACLNet*~\cite{wang2019revisiting} &  0.456 & 0.683 & 0.852 &  1.84 &  0.378  \\  
 & MEP & 0.438 & 0.675 & 0.845 & 1.73 & 0.342  \\ \hline
 \multirow{4}{*}{\rotatebox[origin=c]{90}{Overall}} & HI & 0.644 & \textbf{0.757} & 0.854 & \textbf{3.41} & 0.514 \\
 & AViNet &  0.655 & 0.744 & 0.901 & 2.55 & 0.516 \\
 & ViNet & \textbf{0.671} & 0.742 & \textbf{0.903} & 2.60 & \textbf{0.533} \\
 & DAVE~\cite{tavakoli2019dave} & 0.545 & 0.726 & 0.881 & 2.45 & 0.449  \\
 & ACLNet*~\cite{wang2019revisiting} & 0.475 &  0.700 & 0.870 & 1.98 & 0.379  \\
 & MEP & 0.403 & 0.662 & 0.844 & 1.59 & 0.326 \\ \hline
\end{tabular}
\end{center}
\label{table:dave_results}
\vspace{-1.5em}
\end{table} 

\paragraph{Audio-Visual Datasets}

The comparison of ViNet and AViNet models with state-of-the-art methods on audio-visual datasets are presented in Table~\ref{table:audio_test_results1}. We also present results on ViNet(NF) baseline model, which is a trained of \emph{DHF1K} dataset and not fine-tuned further on audio-visual datasets. The ViNet, ViNet(NF), ACLNet and TASED-Net models are trained without using any audio information. STAViS and AViNet models make use of the audio modality, both during training and inference.  AViNet(B) and AViNet(C) present the two fusion methodologies discussed above \ie  concatenation and bilinear fusion respectively. 

The ViNet model significantly outperforms STAViS on most datasets across most metrics. Surprisingly, the ViNet(NF) model is already competitive, indicating that models trained on \emph{DHF1K} can generalize well to other datasets. Moreover, the results clearly suggest that the visual-only modality, when exploited well, is able to recover most of the underlying performance on the current datasets, as compared to existing state-of-the-art models. The improvements obtained by AViNet(B) and AViNet(C) models over ViNet are marginal at best (with an exception on \emph{Coutrot2} dataset, which is captured in highly specific settings). Hence, in contrast to previous works~\cite{tsiami2020stavis,tavakoli2019dave}, our experimental results do not indicate any clear benefit of incorporating audio in the prediction pipeline. 

We also evaluate the performance of our models on the \emph{AVE} dataset~\cite{tavakoli2019dave}. Although the \emph{AVE} dataset is formed using the sequences from \emph{DIEM}, \emph{Coutrot1}, and \emph{Coutrot2} datasets, it is an interesting dataset because (a) it provides a human upper bound and a lower bound using dataset biases and (b) it provides video level categorization. The upper bound is named Human Infinite (HI) and is computed by splitting the eye-movements of observers into two groups and assessing one group against the other (human vs. human performance). The lower bound is called the Mean Eye Position map (MEP) and is computed from the training sequences. It depicts the center-bias that a model may learn by training on the dataset. It is, hence, a robust lower-bound baseline.

ViNet model outperforms the state-of-the-art approaches on the \emph{AVE} dataset by a significant margin, resonating with the observations on other datasets. Notably, ViNet is able to cross the HI upper bound on AUC-J, CC, and SIM metrics. We further provide a category-wise analysis of both our models on this dataset. It is evident from Table~\ref{table:dave_results} ViNet and AViNet give fairly similar performance across all three categories, giving solid gains over other methods.  

\subsection{The Impact of Audio}
\begin{table}[]
\caption{CC and SIM metrics for AViNet(B), AViNet(C) and STAViS with and without audio(sending zeros for audio) on \emph{Coutrot2} dataset. The predictions are nearly identical as reflected in the metrics.  }
\begin{center}
\begin{tabular}{@{}lll@{}}
\toprule
Method    & CC & SIM \\ \midrule
AViNet(B) &  0.9977  & 0.9979    \\
AViNet(C) & 0.9978   & 0.9990    \\
STAViS &  0.9980  & 0.9981      \\ \bottomrule
\end{tabular}
\end{center}
\label{table:with_without_audio}
\vspace{-1.5em}
\end{table}

We conduct a simple experiment to investigate the role of audio in AViNet and STAViS models. We compare the output predictions obtained with original audio and by sending zeroed-out vector as audio (indicating the absence of audio). To our surprise, the network's prediction maps obtained with and without audio are nearly identical (as presented in Table~\ref{table:with_without_audio}). A qualitative example is shown in Fig. ~\ref{fig:avinet_results}. This indicates that the network learns to be agnostic to audio and gives the same output irrespective of the audio input (zero vector, corresponding audio, or random audio). In summary, the current state-of-the-art audio-saliency models end up learning a visual-only model and that also explains the marginal differences with ViNet and AViNet models in our results (Table~\ref{table:audio_test_results1}). Such marginal differences might arise due to different instances of training or possibly due to a slight variation in the number of parameters. A deeper exploration is left for future work. Finally, the observations contrast with cognitive studies, which suggest clear differences in human gaze patterns when the videos are watched with or without audio~\cite{coutrot2012influence}. The findings open up an interesting avenue for future research for designing architectures that can make better use of the aural modality.

\section{Conclusion}
We propose ViNet, a novel spatio-temporal visual-only architecture that efficiently addresses the problem of saliency detection in videos. We also explored incorporating audio for the task with AViNet by the addition of an auditory module to ViNet. We explore two different fusion techniques for combining audio-visual cues. We perform a comprehensive analysis of both models on 10 different datasets (3 visual and 7 audio-visual). Our models brings significantly gains over the state-of-the-art models. We find that audio does not seem to be playing a major role in audio-visual saliency prediction, even in models that explicitly incorporate audio. Our findings clearly illustrate the need for further explorations in this direction, leading to better models as well as curating datasets which can better utilize the auditory modality. 

\bibliographystyle{IEEEtran}
\bibliography{IEEEabrv,IEEEexample}

\end{document}